\documentclass[english]{article}
\usepackage[T1]{fontenc}
\usepackage[latin9]{inputenc}
\usepackage{geometry}
\usepackage{array}
\usepackage{booktabs}
\usepackage{amsmath}
\usepackage{amssymb}

\makeatletter

\providecommand{\tabularnewline}{\\}

\DeclareMathAlphabet{\mathbfsf}{\encodingdefault}{\sfdefault}{bx}{n}
\newcommand{\upgreektemplate}[2]{#2{
\renewcommand{\alpha}{\upalpha}
\renewcommand{\beta}{\upbeta}
\renewcommand{\theta}{\uptheta}
\renewcommand{\gamma}{\upgamma}
\renewcommand{\lambda}{\uplambda}
\renewcommand{\delta}{\updelta}
\renewcommand{\phi}{\upphi}
\renewcommand{\zeta}{\upzeta}
\renewcommand{\Lambda}{\Uplambda}
\renewcommand{\Gamma}{\Upgamma}
\renewcommand{\Delta}{\Updelta}
\renewcommand{\Theta}{\Uptheta}
#1
}}
\newcommand{\upgreek}[1]{\upgreektemplate{#1}{\mathsf}}
\newcommand{\bupgreek}[1]{\upgreektemplate{#1}{\mathbfsf}}
\usepackage{upgreek}

\usepackage{pdflscape}

\usepackage{hyperref}
\usepackage{xcolor}
\definecolor{mycol}{rgb}{0,0,0.65}
\hypersetup{
    colorlinks,
    linkcolor={mycol},
    citecolor={mycol},
    urlcolor={mycol}
}

\usepackage[nameinlink,capitalize]{cleveref}
\AtBeginDocument{%
\let\ref\cref
}

\makeatother

\usepackage{babel}
\begin{document}
\global\long\def\argmin{\operatornamewithlimits{argmin}}%

\global\long\def\argmax{\operatornamewithlimits{argmax}}%

\global\long\def\prox{\operatornamewithlimits{prox}}%

\global\long\def\diag{\operatorname{diag}}%

\global\long\def\lse{\operatorname{lse}}%

\global\long\def\R{\mathbb{R}}%

\global\long\def\E{\operatornamewithlimits{\mathbb{E}}}%

\global\long\def\P{\operatornamewithlimits{\mathbb{P}}}%

\global\long\def\V{\operatornamewithlimits{\mathbb{V}}}%

\global\long\def\N{\mathcal{N}}%

\global\long\def\L{\mathcal{L}}%

\global\long\def\C{\mathbb{C}}%

\global\long\def\tr{\operatorname{tr}}%

\global\long\def\norm#1{\left\Vert #1\right\Vert }%

\global\long\def\norms#1{\left\Vert #1\right\Vert ^{2}}%

\global\long\def\Dtru{D_{\text{true}}}%

\global\long\def\ehat#1{\underset{#1}{\hat{\mathbb{E}}}}%

\global\long\def\pars#1{\left(#1\right)}%

\global\long\def\pp#1{(#1)}%

\global\long\def\bracs#1{\left[#1\right]}%

\global\long\def\bb#1{[#1]}%

\global\long\def\verts#1{\left\vert #1\right\vert }%

\global\long\def\Verts#1{\left\Vert #1\right\Vert }%

\global\long\def\angs#1{\left\langle #1\right\rangle }%

\global\long\def\KL#1{[#1]}%

\global\long\def\KL#1#2{KL\pars{#1\middle\Vert#2}}%

\global\long\def\div{\text{div}}%

\global\long\def\erf{\text{erf}}%

\global\long\def\vvec{\text{vec}}%

\global\long\def\b#1{\bm{#1}}%

\global\long\def\r#1{\upgreek{#1}}%

\global\long\def\br#1{\bupgreek{\bm{#1}}}%
 
\title{Moment-Matching Conditions for Exponential Families with Conditioning
or Hidden Data}
\author{Justin Domke}
\date{{}}

\maketitle

\begin{abstract}
Maximum likelihood learning with exponential families leads to moment-matching
of the sufficient statistics, a classic result. This can be generalized
to conditional exponential families and/or when there are hidden data.
This document gives a first-principles explanation of these generalized
moment-matching conditions, along with a self-contained derivation.
\end{abstract}

\section{Overview}

This document gives a self-contained introduction to maximum likelihood
learning in normal or conditional exponential families, with our without
hidden variables. The purpose is to show that the standard ``moment
matching'' property extends to all of these cases, and that there
is a mnemonic that makes it easy to remember all cases. All the results
are summarized in \ref{fig:summary} -- the rest of this document
derives these results and gives some discussion.

While self-contained, this document is not intended as a first or
comprehensive introduction to exponential families \cite{Lehmann_1998_Theorypointestimation,Bickel_2001_Mathematicalstatisticsbasic,Barndorff-Nielsen_2014_InformationExponentialFamilies}.
There is little motivation and no examples. Rather, this is intended
for a reader familiar with typical exponential families who wishes
to understand how maximum likelihood works with hidden variables and/or
conditioning.

\subsection{Notation}

We use sans-serif characters (e.g. $\r x$) to refer to random variables,
and roman characters (e.g. $x$) to refer to specific values. Given
a finite dataset, $\hat{\E}$ denotes empirical expectations. For
example, given a dataset $x^{(1)},x^{(2)},\cdots,x^{(n)},$ $\ehat{\r x}f\pp{\r x}=\frac{1}{n}\sum_{i=1}^{n}f\pars{x^{(i)}}.$

\section{Exponential Family\label{sec:Exponential-Family}}

The material in is section is all classic. We define an exponential
family as
\begin{eqnarray*}
p_{\theta}\pp x & = & h\pp x\exp\pars{\theta^{\top}T(x)-A(\theta)},\\
A\pp{\theta} & = & \log\sum_{x}h\pp x\exp\pars{\theta^{\top}T\pp x}.
\end{eqnarray*}
The objects are:
\begin{itemize}
\item $x$ - the variable. For simplicity, we assume $x$ takes values in
a finite set, though the properties extend easily to the case where
$x$ is continuous.
\item $T\pp x$ - the ``sufficient statistics''. Abstractly, this is some
vector-valued function that extracts important ``features'' of $x$.
\item $\theta$ - parameter vector.
\item $h\pp x$ - the scaling constant. This will often just be one.
\item $A\pp{\theta}$ - the ``cumulant'' or ``log-partition'' function,
which makes $p_{\theta}$ sum to one.
\end{itemize}
\textbf{Gradient of log-partition function.} The key property that
gives maximum-likelihood its elegance with exponential families is
the gradient of $A$. This gradient is the expected value of the sufficient
statistics, under the current parameters $\theta$. This is easy to
show: 
\begin{eqnarray*}
\frac{dA\pp{\theta}}{d\theta} & = & \frac{d}{d\theta}\log\sum_{x}h\pp x\exp\pars{\theta^{\top}T\pp x}\\
 & = & \frac{\frac{d}{d\theta}\sum_{x}h\pp x\exp\pars{\theta^{\top}T\pp x}}{\sum_{x}h\pp x\exp\pars{\theta^{\top}T\pp x}}\\
 & = & \frac{\sum_{x}h\pp x\exp\pars{\theta^{\top}T\pp x}T\pp x}{\exp\pp{A\pp{\theta}}}\\
 & = & \sum_{x}h\pp x\exp\pars{\theta^{\top}T\pp x-A\pp{\theta}}T\pp x\\
 & = & \sum_{x}p_{\theta}\pp xT\pp x\\
 & = & \E_{p_{\theta}\pp{\r x}}\bb{T\pp{\r x}}
\end{eqnarray*}
\textbf{Objective for a single datum.} Suppose $x$ is a single datum.
The log-likelihood of that datum is clearly
\[
\log p_{\theta}\pp x=\log h\pp x+\theta^{\top}T\pp x-A\pp{\theta}.
\]

\textbf{Form of learning objective}. Take a dataset $x^{(1)},x^{(2)},\cdots,x^{(n)}.$
In this case, it makes sense to maximize the mean log-likelihood
\begin{eqnarray*}
L\pp{\theta} & = & \frac{1}{n}\sum_{i=1}^{n}\log p_{\theta}\pars{x^{(i)}}.
\end{eqnarray*}

\textbf{Likelihood of a dataset}. Substituting the form of $p_{\theta}$,
the learning objective becomes 
\begin{eqnarray*}
L\pp{\theta} & = & \frac{1}{n}\sum_{i=1}^{n}\log p_{\theta}\pars{x^{(i)}}\\
 & = & \frac{1}{n}\sum_{i=1}^{n}\pars{\log h\pp{x^{(i)}}+\theta^{\top}T\pp{x^{(i)}}-A\pp{\theta}}
\end{eqnarray*}

\textbf{Alternate form for the likelihood of a dataset.} It is informative
to write the learning objective using \emph{empirical expectations}
rather than explicit sums. If we do this, $L$ can be re-written as
\begin{eqnarray*}
L\pp{\theta} & = & \ehat{\r x}\bracs{\log h\pp{\r x}+\theta^{\top}T\pp{\r x}}-A\pp{\theta}.
\end{eqnarray*}

\textbf{Condition at optimum}. Now, suppose that one has found $\theta$
that maximimizes $L$. Then, it must be true that the gradient of
$L$ is zero. Thus, it must be true that
\begin{eqnarray*}
0 & = & \frac{dL\pp{\theta}}{d\theta}\\
 & = & \frac{d}{d\theta}\ehat{\r x}\bracs{\log h\pp{\r x}+\theta^{\top}T\pp{\r x}}-\frac{d}{d\theta}A\pp{\theta}\\
 & = & \ehat{\r x}\bracs{T\pp{\r x}}-\E_{p_{\theta}\pp{\r x}}\bb{T\pp{\r x}}.
\end{eqnarray*}
Thus, if $\theta$ are the maximum-likelihood parameters, it must
be true that
\begin{equation}
\ehat{\r x}\bracs{T\pp{\r x}}=\E_{p_{\theta}\pp{\r x}}\bb{T\pp{\r x}}.\label{eq:moment-matching-conditions}
\end{equation}
These are the classic ``moment-matching'' conditions of maximum-likelihood.

\section{Conditional Exponential Family}

A conditional exponential family can be written as

\begin{eqnarray}
p_{\theta}\pp{y|x} & = & h\pp{x,y}\exp\pars{\theta^{\top}T(x,y)-A(\theta)},\label{eq:conditional-efam-definition}\\
A(x,\theta) & = & \log\sum_{y}h(x,y)\exp\theta^{\top}T(x,y).\nonumber 
\end{eqnarray}

One could derive this by defining a joint exponential family over
the joint space $\pp{x,y}$ and then conditioning it. However, it
is best to think of it as a new model definition. This emphasizes
that $p_{\theta}\pp x$ is not even defined.

\textbf{Gradient of log-partition function.} In conditional exponential
families, the gradient of $A$ becomes the \emph{conditional} expected
value of the sufficient statistics, under the current parameters $\theta$.

\begin{eqnarray}
\frac{dA\pp{x,\theta}}{d\theta} & = & \frac{d}{d\theta}\log\sum_{y}h\pp{x,y}\exp\pars{\theta^{\top}T\pp{x,y}}\nonumber \\
 & = & \frac{\frac{d}{d\theta}\sum_{y}h\pp{x,y}\exp\pars{\theta^{\top}T\pp{x,y}}}{\sum_{y}h\pp{x,y}\exp\pars{\theta^{\top}T\pp{x,y}}}\nonumber \\
 & = & \frac{\sum_{y}h\pp{x,y}\exp\pars{\theta^{\top}T\pp{x,y}}T\pp{x,y}}{\exp\pp{A\pp{x,\theta}}}\nonumber \\
 & = & \sum_{y}h\pp{x,y}\exp\pars{\theta^{\top}T\pp x-A\pp{x,\theta}}T\pp{x,y}\nonumber \\
 & = & \sum_{y}p_{\theta}\pp{y|x}T\pp{x,y}\nonumber \\
 & = & \E_{p_{\theta}\pp{\r y|x}}\bb{T\pp{x,\r y}}\label{eq:conditional-efam-gradient}
\end{eqnarray}
\textbf{Objective for a single datum.} Suppose $\pp{x,y}$ is a single
datum. Since $p_{\theta}\pp x$ is not defined, it doesn't even make
sense to talk about $p_{\theta}\pp{x,y}$. Instead, the natural object
is the conditional log-likelihood. This is

\[
\log p_{\theta}\pp{y|x}=\log h\pp{x,y}+\theta^{\top}T\pp{x,y}-A\pp{x,\theta}.
\]

\textbf{Form of learning objective}. Take a dataset $\pp{x^{(1)},y^{(1)}},\cdots,\pp{x^{(n)},y^{(n)}}.$
The mean conditional log-likelihood is
\begin{eqnarray*}
L\pp{\theta} & = & \frac{1}{n}\sum_{i=1}^{n}\log p_{\theta}\pars{y^{(i)}\vert x^{(i)}}.
\end{eqnarray*}

\textbf{Likelihood of a dataset}. Substituting the form of $p_{\theta}$,
the learning objective becomes 
\begin{eqnarray*}
L\pp{\theta} & = & \frac{1}{n}\sum_{i=1}^{n}\log p_{\theta}\pars{y^{(i)}\vert x^{(i)}}\\
 & = & \frac{1}{n}\sum_{i=1}^{n}\pars{\log h\pp{x^{(i)},y^{(i)}}+\theta^{\top}T\pp{x^{(i)},y^{(i)}}-A\pp{x^{(i)},\theta}}.
\end{eqnarray*}

\textbf{Alternate form for the likelihood of a dataset.} Re-written
in terms of empirical expectations, the learning objective is 
\begin{eqnarray*}
L\pp{\theta} & = & \ehat{\r x,\r y}\bracs{\log h\pp{\r x,\r y}+\theta^{\top}T\pp{\r x,\r y}-A\pp{\r x}}.
\end{eqnarray*}

\textbf{Condition at optimum}. Now, suppose that one has found $\theta$
that maximimizes $L$. Then, it must be true that the gradient of
$L$ is zero. Thus, it must be true that
\begin{eqnarray*}
0 & = & \frac{dL\pp{\theta}}{d\theta}\\
 & = & \frac{d}{d\theta}\ehat{\r x}\bracs{\log h\pp{\r x,\r y}+\theta^{\top}T\pp{\r x,\r y}-A\pp{\r x,\theta}}\\
 & = & \ehat{\r x}\bracs{T\pp{\r x,\r y}}-\frac{d}{d\theta}\ehat{\r x}\bracs{A\pp{\r x,\theta}}\\
 & = & \ehat{\r x}\bracs{T\pp{\r x,\r y}}-\ehat{\r x}\E_{p_{\theta}\pp{\r y|x}}\bb{T\pp{\r x,\r y}}.
\end{eqnarray*}
Thus, if $\theta$ are the maximum-likelihood parameters, it must
be true that
\begin{equation}
\ehat{\r x,\r y}\bracs{T\pp{\r x,\r y}}=\ehat{\r x}\E_{p_{\theta}\pp{\r y|x}}\bb{T\pp{\r x,\r y}}.\label{eq:moment-matching-conditional}
\end{equation}
This generalizes the moment-matching conditions from \ref{eq:moment-matching-conditions}.
Intuitively, we still have a ``data expectation on one side'' and
a ``model expectation on the other side''. However, the model does
not define a distribution over $x$. Intuitively, \ref{eq:moment-matching-conditional}
``fills in'' the model expectaion on the right-hand side using the
data.

\section{Exponential Family with Hidden Variables}

Take an exponential family jointly over $\pp{x,u}$

\begin{eqnarray*}
p_{\theta}\pp{x,u} & = & h(x,u)\exp\pars{\theta^{\top}T(x,u)-A(\theta)}\\
A(\theta) & = & \log\sum_{x,u}h(x,u)\exp\theta^{\top}T(x,u)
\end{eqnarray*}

\textbf{Marginal distribution}. What is the marginal distribution
of $p_{\theta}$ over $x$? It's not hard to show that
\begin{eqnarray*}
p_{\theta}(x) & = & \sum_{u}p_{\theta}\pp{x,u}\\
 & = & \sum_{u}h(x,u)\exp\pars{\theta^{\top}T(x,u)-A(\theta)}\\
 & = & \exp\pars{\log\pars{\sum_{u}h(x,u)\exp\pars{\theta^{\top}T(x,u)-A(\theta)}}}\\
 & = & \exp\pars{\log\pars{\sum_{u}h(x,u)\exp\pars{\theta^{\top}T(x,u)}}-A(\theta)}\\
 & = & \exp\pars{A\pp{x,\theta}-A(\theta)},
\end{eqnarray*}
where we define
\[
A\pp{x,\theta}=\log\sum_{u}h(x,u)\exp\pars{\theta^{\top}T(x,u)}.
\]
One should think of the extra argument of $x$ in $A\pp{x,\theta}$
as meaning that $x$ remains fixed in the sum defining this new log-partition
function.

\textbf{Gradient of log-partition function.} By the same logic as
in \ref{sec:Exponential-Family} the gradient of $A\pp{\theta}$ is
\[
\frac{dA(\theta)}{d\theta}=\E_{p_{\theta}(\r x,\r u)}\bracs{T(\r x,\r u)}.
\]

Meanwhile, the gradient of $A\pp{x,\theta}$ is
\begin{eqnarray*}
\frac{dA(x,\theta)}{d\theta} & = & \frac{d}{d\theta}\log\sum_{u}h(x,u)\exp\pars{\theta^{\top}T(x,u)}\\
 & = & \frac{\frac{d}{d\theta}\sum_{u}h(x,u)\exp\pars{\theta^{\top}T(x,u)}}{\sum_{u}h(x,u)\exp\pars{\theta^{\top}T(x,u)}}\\
 & = & \frac{\sum_{u}h(x,u)\exp\pars{\theta^{\top}T(x,u)}T\pp{x,u}}{\exp\pp{A\pp{x,\theta}}}\\
 & = & \sum_{u}h(x,u)\exp\pars{\theta^{\top}T(x,u)-A\pp{x,\theta}}T\pp{x,u}\\
 & = & \E_{p_{\theta}(\r u|x)}\bracs{T(x,\r u)}
\end{eqnarray*}
Notice here that $x$ is a fixed value, not a random variable.

\textbf{Objective for a single datum.} Suppose $x$ is a single datum,
with the corresponding $u$ left unobserved. It is impossible to evaluate
$p_{\theta}\pp{x,u}.$ A natural alternative is the marginal log-likelihood.
This is

\[
\log p_{\theta}\pp x=A\pp{x,\theta}-A(\theta).
\]

\textbf{Form of learning objective}. Take a dataset $\pp{x^{(1)}},\cdots,\pp{x^{(n)}}.$
The mean marginal log-likelihood is
\begin{eqnarray*}
L\pp{\theta} & = & \frac{1}{n}\sum_{i=1}^{n}\log p_{\theta}\pars{x^{(i)}}.
\end{eqnarray*}

\textbf{Likelihood of a dataset}. Substituting the form of $p_{\theta}\pp x$,
the learning objective becomes 
\begin{eqnarray*}
L\pp{\theta} & = & \frac{1}{n}\sum_{i=1}^{n}\log p_{\theta}\pars{x^{(i)}}\\
 & = & \frac{1}{n}\sum_{i=1}^{n}\pars{A\pp{x^{(i)},\theta}-A\pp{\theta}}.
\end{eqnarray*}

\textbf{Alternate form for the likelihood of a dataset.} Re-written
in terms of empirical expectations, the learning objective is 
\begin{eqnarray*}
L\pp{\theta} & = & \ehat{\r x}\bracs{A\pp{\r x,\theta}-A\pp{\r x}}.
\end{eqnarray*}

\textbf{Condition at optimum}. Now, suppose that one has found $\theta$
that maximimizes $L$. Then, it must be true that the gradient of
$L$ is zero. Thus, it must be true that
\begin{eqnarray*}
0 & = & \frac{dL\pp{\theta}}{d\theta}\\
 & = & \frac{d}{d\theta}\ehat{\r x}\bracs{A\pp{\r x,\theta}-A\pp{\theta}}\\
 & = & \ehat{\r x}\bracs{\frac{d}{d\theta}A\pp{\r x,\theta}}-\frac{d}{d\theta}A\pp{\theta}\\
 & = & \ehat{\r x}\E_{p_{\theta}(\r u|\r x)}\bracs{T(\r x,\r u)}-\E_{p_{\theta}(\r x,\r u)}\bracs{T(\r x,\r u)}.
\end{eqnarray*}
Thus, if $\theta$ are the maximum-likelihood parameters, it must
be true that
\begin{equation}
\ehat{\r x}\E_{p_{\theta}(\r u|\r x)}\bracs{T(\r x,\r u)}=\E_{p_{\theta}(\r x,\r u)}\bracs{T(\r x,\r u)}.\label{eq:moment-matching-hidden}
\end{equation}

The again generalizes the moment-matching conditions from \ref{eq:moment-matching-conditions}.
Intuitively, we still have a ``data expectation on one side'' and
a ``model expectation on the other side''. However, the data does
not contain values for $u$. Intuitively, \ref{eq:moment-matching-hidden}
``fills in'' the data expectation on the left-hand side using the
model.

This is in constrast to the conditions for the conditional exponential
family seen in \ref{eq:moment-matching-conditional}. There, it was
the \emph{model} that was missing data, which was ``filled in''
by the data on the right-hand side.

\section{Conditional Exponential Family with Hidden Variables}

Take an exponential family jointly over $\pp{y,u}$ but conditional
on $x$. This is essentially the same as \ref{eq:conditional-efam-definition},
just with $y$ transformed to $\pp{y,u}.$

\begin{eqnarray*}
p_{\theta}(y,u\vert x) & = & h(x,y,u)\exp\pars{\theta^{\top}T(x,y,u)-A(x,\theta)}\\
A(x,\theta) & = & \log\sum_{y,u}h(x,y,u)\exp\theta^{\top}T(x,y,u).
\end{eqnarray*}

\textbf{Marginal distribution}. What is the marginal distribution
of $p_{\theta}$ over $y$ given $x$? This is

\begin{eqnarray*}
p_{\theta}(y\vert x) & = & \sum_{u}p_{\theta}\pp{y,u|x}\\
 & = & \sum_{u}h(x,y,u)\exp\pars{\theta^{\top}T(x,y,u)-A(x,\theta)}\\
 & = & \exp\pars{\log\pars{\sum_{u}h(x,y,u)\exp\pars{\theta^{\top}T(x,y,u)-A(x,\theta)}}}\\
 & = & \exp\pars{\log\pars{\sum_{u}h(x,y,u)\exp\pars{\theta^{\top}T(x,y,u)}}-A(x,\theta)}\\
 & = & \exp\pars{A\pp{x,y,\theta}-A(x,\theta)},
\end{eqnarray*}
where we define
\[
A\pp{x,\theta}=\log\pars{\sum_{u}h(x,y,u)\exp\pars{\theta^{\top}T(x,y,u)}}.
\]
One should think of the extra argument of $x$ in $A\pp{x,\theta}$
as meaning that $x$ remains fixed in the sum defining this new log-partition
function.

\textbf{Conditional distribution}. We will also need the conditional
$p_{\theta}\pp{u|y,x}$. This is
\begin{eqnarray*}
p_{\theta}\pp{u|y,x} & = & \frac{p_{\theta}\pp{y,u|x}}{p_{\theta}\pp{y|x}}\\
 & = & \frac{h(x,y,u)\exp\pars{\theta^{\top}T(x,y,u)-A(x,\theta)}}{\exp\pars{A\pp{x,y,\theta}-A(x,\theta)}}\\
 & = & h(x,y,u)\exp\pars{\theta^{\top}T(x,y,u)-A\pp{x,y,\theta}}.
\end{eqnarray*}
(This formula does not appear in the cheatsheet at the start of this
document.)

\textbf{Gradient of log-partition function.} By the same logic as
in \ref{eq:conditional-efam-gradient} the gradient of $A\pp{\theta}$
is 
\[
\frac{dA(x,\theta)}{d\theta}=\E_{p_{\theta}(\r y,\r u\vert x)}\bracs{T(x,\r y,\r u)}.
\]

Meanwhile, the gradient of $A\pp{x,\theta}$ is

\begin{eqnarray*}
\frac{dA(x,y,\theta)}{d\theta} & = & \frac{d}{d\theta}\log\pars{\sum_{u}h(x,y,u)\exp\pars{\theta^{\top}T(x,y,u)}}\\
 & = & \frac{\frac{d}{d\theta}\sum_{u}h(x,y,u)\exp\pars{\theta^{\top}T(x,y,u)}}{\sum_{u}h(x,y,u)\exp\pars{\theta^{\top}T(x,y,u)}}\\
 & = & \frac{\sum_{u}h(x,y,u)\exp\pars{\theta^{\top}T(x,y,u)}T(x,y,u)}{\exp\pars{A\pp{x,y,\theta}}}\\
 & = & \sum_{u}h(x,y,u)\exp\pars{\theta^{\top}T(x,y,u)-A\pp{x,y,\theta}}T(x,y,u)\\
 & = & \sum_{u}p_{\theta}\pp{u|y,x}T(x,y,u)\\
 & = & \E_{p_{\theta}(\r u\vert x,y)}\bracs{T(x,y,\r u)}
\end{eqnarray*}
Notice here that $x$ and $y$ a fixed values.

\textbf{Objective for a single datum.} Suppose $\pp{x,y}$ is a single
datum, with the corresponding $u$ left unobserved. The natural objective
is the ``marginal conditional'' log likelihood $p_{\theta}\pp{y|x}$.
($u$ is marginalized out, while $x$ is conditioned upon.) This objective
is

\[
\log p_{\theta}\pp{y|x}=A\pp{x,y,\theta}-A(x,\theta).
\]

\textbf{Form of learning objective}. Take a dataset $\pp{x^{(1)},y^{(1)}},\cdots,\pp{x^{(n)},y^{(1)}}.$
The mean marginal conditional log-likelihood is
\begin{eqnarray*}
L\pp{\theta} & = & \frac{1}{n}\sum_{i=1}^{n}\log p_{\theta}\pars{y^{(i)}\vert x^{(i)}}.
\end{eqnarray*}

\textbf{Likelihood of a dataset}. Substituting the form of $p_{\theta}\pp x$,
the learning objective becomes 
\begin{eqnarray*}
L\pp{\theta} & = & \frac{1}{n}\sum_{i=1}^{n}\log p_{\theta}\pars{y^{(i)}\vert x^{(i)}}\\
 & = & \frac{1}{n}\sum_{i=1}^{n}\pars{A\pp{x^{(i)},y^{(i)},\theta}-A(x^{(i)},\theta)}.
\end{eqnarray*}

\textbf{Alternate form for the likelihood of a dataset.} Re-written
in terms of empirical expectations, the learning objective is 
\begin{eqnarray*}
L\pp{\theta} & = & \ehat{\r x}\bracs{A\pp{\r x,\r y,\theta}-A(\r x,\theta)}.
\end{eqnarray*}

\textbf{Condition at optimum}. Now, suppose that one has found $\theta$
that maximizes $L$. Then, it must be true that the gradient of $L$
is zero. Thus, it must be true that
\begin{eqnarray*}
0 & = & \frac{dL\pp{\theta}}{d\theta}\\
 & = & \frac{d}{d\theta}\ehat{\r x,\r y}\bracs{A\pp{\r x,\r y,\theta}-A(\r x,\theta)}\\
 & = & \ehat{\r x,\r y}\bracs{\frac{d}{d\theta}A\pp{\r x,\r y,\theta}}-\ehat{\r x}\bracs{\frac{d}{d\theta}A(\r x,\theta)}\\
 & = & \ehat{\r x,\r y}\E_{p_{\theta}(\r u\vert\r x,\r y)}\bracs{T(\r x,\r y,\r u)}-\ehat{\r x}\E_{p_{\theta}(\r y,\r u\vert x)}\bracs{T(\r x,\r y,\r u)}.
\end{eqnarray*}
Thus, if $\theta$ are the maximum-likelihood parameters, it must
be true that
\begin{equation}
\ehat{\r x,\r y}\E_{p_{\theta}(\r u\vert\r x,\r y)}\bracs{T(\r x,\r y,\r u)}=\ehat{\r x}\E_{p_{\theta}(\r y,\r u\vert x)}\bracs{T(\r x,\r y,\r u)}.\label{eq:moment-matching-conditional-hidden}
\end{equation}

The again generalizes the moment-matching conditions from \ref{eq:moment-matching-conditions},
but in both of the ways that \ref{eq:moment-matching-conditional}
and \ref{eq:moment-matching-hidden} did. Intuitively, we still have
a ``data expectation on one side'' and a ``model expectation on
the other side''. However, the data does not contain values for $u$.
Intuitively, \ref{eq:moment-matching-conditional-hidden} ``fills
in'' the data expectation on the left-hand side using the model.
Moreover, the model does not define $p_{\theta}\pp x$. The data ``fills
in'' that on the right-hand side.

\section{Mnemonic}

We propose the following mnemonic to remember the final moment matching
optimality conditions:
\begin{itemize}
\item Put ``as much data expectation'' $\hat{\E}\bb{T\pp{\cdot}}$ on
the left hand side as possible.
\item Put ``as much model expectation'' $\E_{p_{\theta}}\bb{T\pp{\cdot}}$
on the right hand side as possible.
\item If variables are ``missing'' from either the model or the data,
use the other to ``fill them in''.
\item At maximum likelihood, the two expectations must be equal.
\end{itemize}
We could also write this (very informally) as

\[
"\underbrace{\ \hat{\E}\ \bracs{\ T\pp{\cdot}\ }\ }_{\text{Data Expectation: fill in with \ensuremath{\E_{p_{\theta}}} as needed}}=\underbrace{\ \E_{p_{\theta}}\ \bracs{\ T\pp{\cdot}\ }\ }_{\text{Model Expectation: fill in with \ensuremath{\hat{\E}} as needed}}".
\]

The idea is to start with this equation, then ``fill in'' the left-hand
side with expectations over $p_{\theta}$ and the right-hand side
with empirical expectations as necessary in order for the condition
to make sense. This is easiest to see through looking at each of the
cases.

\subsection{Exponential Family}

For a normal exponential family $p_{\theta}\pp x$ and standard likelihood
with a dataset $x^{(1)}\cdots x^{(n)}$ the heuristic gives
\[
\hat{\E}_{\r x}\ \bb{\ T\pp{\r x}\ }=\E_{p_{\theta}\pp{\r x}}\ \bb{\ T\pp{\r x}\ }.
\]
In this case, no further changes are needed.

\subsection{Conditional Exponential Family}

For a conditional exponential family $p_{\theta}\pp{y|x}$ and conditional
likelihood with a dataset $\pp{x^{(1)},y^{(1)}}\cdots\pp{x^{(n)},y^{(n)}}$
the heuristic gives:
\[
"\hat{\E}_{\r x,\r y}\ \bb{\ T\pp{\r x,\r y}\ }=\E_{p_{\theta}\pp{\r x,\r y}}\ \bb{\ T\pp{\r x,\r y}\ }."
\]

This is in quotes because it is not correct and is not a real equation.
The problem is that the right-hand side is meaningless since $p_{\theta}$
does not define an expectation over $\r x$. Instead, we must fill
in with the data. This gives the correct condition
\[
\hat{\E}_{\r x,\r y}\ \bb{\ T\pp{\r x,\r y}\ }=\ehat{\r x}\E_{p_{\theta}\pp{\r y|\r x}}\ \bb{\ T\pp{\r x,\r y}\ }.
\]

\subsection{Exponential Family with Hidden Variables}

Now, take an exponential family $p_{\theta}\pp{x,u}$ that is over
$x$ and $u$ together but where $u$ is hidden in the dataset $x^{(1)},\cdots,x^{(n)}$.
If maximizing the marginal likelihood over $x$, then, the heuristic
gives the condition
\[
"\hat{\E}_{\r x,\r u}\ \bb{\ T\pp{\r x,\r u}\ }=\E_{p_{\theta}\pp{\r x,\r u}}\ \bb{\ T\pp{\r x,\r u}\ }."
\]
Again, this is in quotes because it is not correct and not a real
equation. The problem now is the the left-hand side is meaningless
since we do not have data for $u$. The solution is to fill in using
the model. This gives the correct condition 
\[
\hat{\E}_{\r x}\E_{p_{\theta}\pp{\r u|\r x}}\ \bb{\ T\pp{\r x,\r u}\ }=\E_{p_{\theta}\pp{\r x,\r u}}\ \bb{\ T\pp{\r x,\r u}\ }.
\]

\subsection{Conditional Exponential Family with Hidden Variables}

Finally take a conditional exponential family $p_{\theta}\pp{y,u|x}$
where $u$is hidden in the dataset $\pp{x^{(1)},y^{(1)}},\cdots,\pp{x^{(n)},y^{(n)}}.$
The objective is maximize the marginal conditional likelihood (marginalizing
out $u$) and conditioning on $x$. The heuristic gives the (again,
not correct) condition
\[
"\hat{\E}_{\r x,\r y,\r u}\ \bb{\ T\pp{\r x,\r y,\r u}\ }=\E_{p_{\theta}\pp{\r x,\r y,\r u}}\ \bb{\ T\pp{\r x,\r y,\r u}\ }."
\]
There are two issues. Firstly, the left-hand side is incorrect because
the data does not provide $u$. Secondly, the right-hand side is incorrect
because the model does not define $p_{\theta}\pp x$. If we fix both
of these, we get the correct condition.
\[
\ehat{\r x,\r y}\E_{p_{\theta}(\r u|\r x,\r y)}\bracs{T(\r x,\r y,\r u)}=\ehat{\r x}\E_{p_{\theta}(\r y,\r u|\r x)}\bracs{T(\r x,\r y,\r u)}.
\]

\section*{Acknowledgements}

I thank Dimitri Semenovich for encouragement and helpful comments.

\bibliographystyle{plain}
\bibliography{justindomke_zotero_betterbibtex}

\newgeometry{left=0.1cm,bottom=0.1cm,right=0.1cm,top=0.1cm}
\begin{landscape}
\vspace{.5cm}
\everymath={\displaystyle}

\renewcommand{\arraystretch}{2.3}
\setlength{\tabcolsep}{1pt}

\vspace{2.5cm}

\begin{figure}
\begin{tabular}{>{\centering}p{0.1\columnwidth}>{\centering}p{0.17\columnwidth}>{\centering}p{0.23\columnwidth}>{\centering}p{0.22\columnwidth}>{\centering}p{0.25\columnwidth}}
 & Exponential Family & Conditional Exponential Family & Exponential Family with Hidden Variables & Conditional Exponential Family with Hidden Variables\tabularnewline
\cmidrule{2-5} \cmidrule{3-5} \cmidrule{4-5} \cmidrule{5-5} 
Definition & {\footnotesize{}$p_{\theta}(x)=h(x)\exp\pars{\theta^{\top}T(x)-A(\theta)}$} & {\footnotesize{}$p_{\theta}(y\vert x)=h(x,y)\exp\pars{\theta^{\top}T(x,y)-A(x,\theta)}$} & {\footnotesize{}$p_{\theta}(x,u)=h(x,u)\exp\pars{\theta^{\top}T(x,u)-A(\theta)}$} & {\footnotesize{}$p_{\theta}(y,u\vert x)=h(x,y,u)\exp\pars{\theta^{\top}T(x,y,u)-A(x,\theta)}$}\tabularnewline
 &  &  & {\footnotesize{}$p_{\theta}(x)=\exp\pars{A(x,\theta)-A(\theta)}$
}\linebreak{}
{\footnotesize{}(consequence of above)} & {\footnotesize{}$p_{\theta}(y\vert x)=\exp\pars{A(y,x,\theta)-A(x,\theta)}$
}\linebreak{}
{\footnotesize{}(consequence of above)}\tabularnewline
Log-partition function & {\footnotesize{}$A(\theta)=\log\sum_{x}h(x)\exp\theta^{\top}T(x)$} & {\footnotesize{}$A(x,\theta)=\log\sum_{y}h(x,y)\exp\theta^{\top}T(x,y)$} & {\footnotesize{}$A(\theta)=\log\sum_{x,u}h(x,u)\exp\theta^{\top}T(x,u)$} & {\footnotesize{}$A(x,\theta)=\log\sum_{y,u}h(x,y,u)\exp\theta^{\top}T(x,y,u)$}\tabularnewline
 &  &  & {\footnotesize{}$A(x,\theta)=\log\sum_{u}h(x,u)\exp\theta^{\top}T(x,u)$} & {\footnotesize{}$A(x,y,\theta)=\log\sum_{u}h(x,y,u)\exp\theta^{\top}T(x,y,u)$}\tabularnewline
Gradient of log-partition function & {\footnotesize{}$\frac{dA(\theta)}{d\theta}=\E_{p_{\theta}(\r x)}\bracs{T(\r x)}$} & {\footnotesize{}$\frac{dA(x,\theta)}{d\theta}=\E_{p_{\theta}(\r y|x)}\bracs{T(x,\r y)}$} & {\footnotesize{}$\frac{dA(\theta)}{d\theta}=\E_{p_{\theta}(\r x,\r u)}\bracs{T(\r x,\r u)}$} & {\footnotesize{}$\frac{dA(x,\theta)}{d\theta}=\E_{p_{\theta}(\r y,\r u\vert x)}\bracs{T(x,\r y,\r u)}$}\tabularnewline
 &  &  & {\footnotesize{}$\frac{dA(x,\theta)}{d\theta}=\E_{p_{\theta}(\r u|x)}\bracs{T(x,\r u)}$} & {\footnotesize{}$\frac{dA(x,y,\theta)}{d\theta}=\E_{p_{\theta}(\r u\vert x,y)}\bracs{T(x,y,\r u)}$}\tabularnewline
Objective for a single datum & {\footnotesize{}$\log p_{\theta}(x)=\log h\pp x+\theta^{\top}T(x)-A(\theta)$} & {\footnotesize{}$\log p_{\theta}(y\vert x)=h\pp{x,y}+\theta^{\top}T(x,y)-A(x,\theta)$} & {\footnotesize{}$\log p_{\theta}(x)=\log h\pp{x,u}+A(x,\theta)-A(\theta)$} & {\footnotesize{}$\log p_{\theta}(y|x)=A(x,y,\theta)-A(x,\theta)$}\tabularnewline
Form of learning objective & {\footnotesize{}$L(\theta)=\frac{1}{n}\sum_{i=1}^{n}\log p(x^{(i)}\vert\theta)$} & {\footnotesize{}$L(\theta)=\frac{1}{n}\sum_{i=1}^{n}\log p_{\theta}(y^{(i)}\vert x^{(i)})$} & {\footnotesize{}$L(\theta)=\frac{1}{N}\sum_{n=1}^{N}\log p_{\theta}(x^{(i)})$} & {\footnotesize{}$L(\theta)=\frac{1}{n}\sum_{i=1}^{n}\log p_{\theta}(y^{(i)}\vert x^{(i)})$}\tabularnewline
Alternative form & {\footnotesize{}$L(\theta)=\frac{1}{n}\sum_{i=1}^{n}\theta^{\top}T(x^{(i)})-A(\theta)$} & {\footnotesize{}$L(\theta)=\frac{1}{n}\sum_{i=1}^{n}\pars{h\pp{x^{(i)},y^{(i)}}+\theta^{\top}T(x^{(i)},y^{(i)})-A(x^{(i)},\theta)}$} & {\footnotesize{}$L(\theta)=\frac{1}{n}\sum_{i=1}^{n}\pars{A(x^{(i)},\theta)-A(\theta)}$} & {\footnotesize{}$L(\theta)=\frac{1}{n}\sum_{i=1}^{n}\pars{A(x^{(i)},y^{(i)},\theta)-A(x^{(i)},\theta)}$}\tabularnewline
Alternative form & {\footnotesize{}$L(\theta)=\ehat{\r x}\bracs{\theta^{\top}T(\r x)}-A(\theta)$} & {\footnotesize{}$L(\theta)=\ehat{\r x,\r y}\bracs{\log h\pp{\r x,\r y}+\theta^{\top}T(\r x,\r y)}-\ehat{\r x}\bracs{A(\r x,\theta)}$} & {\footnotesize{}$L(\theta)=\ehat{\r x}\bracs{A(\r x,\theta)}-A(\theta)$} & {\footnotesize{}$L(\theta)=\ehat{\r x,\r y}A(\r x,\r y,\theta)-\ehat{\r x}A(\r x,\theta)$}\tabularnewline
Condition at optimum & {\footnotesize{}$\ehat{\r x}[T(\r x)]=\E_{p_{\theta}(\r x)}\bracs{T(\r x)}$} & {\footnotesize{}$\ehat{\r x,\r y}[T(\r x,\r y)]=\ehat{\r x}\E_{p_{\theta}(\r y|\r x)}\bracs{T(\r x,\r y)}$} & {\footnotesize{}$\ehat{\r x}\E_{p_{\theta}(\r u\vert\r x)}\bracs{T(\r x,\r u)}=\E_{p_{\theta}(\r x,\r u)}\bracs{T(\r x,\r u)}$} & {\footnotesize{}$\ehat{\r x,\r y}\E_{p_{\theta}(\r u|\r x,\r y)}\bracs{T(\r x,\r y,\r u)}=\ehat{\r x}\E_{p_{\theta}(\r y,\r u|\r x)}\bracs{T(\r x,\r y,\r u)}$}\tabularnewline
Note &  & {\footnotesize{}$p_{\theta}(x)$ undefined \& irrelevant} & {\footnotesize{}$p_{\theta}(u\vert x)$ defined \& relevant} & {\footnotesize{}$p_{\theta}(x)$ undefined \& irrelevant}\linebreak{}
{\footnotesize{} $p_{\theta}(u\vert x,y)$ defined and relevant}\tabularnewline
\end{tabular}

\caption{A summary of all moment-matching conditions.\label{fig:summary}}
\end{figure}

\end{landscape}
\restoregeometry
\end{document}